\documentclass[10pt,twocolumn]{article}

\usepackage[margin=0.75in]{geometry}
\usepackage{amsmath,amssymb,amsfonts}
\usepackage{graphicx}
\usepackage{booktabs}
\usepackage{cite}
\usepackage{hyperref}
\usepackage{caption}
\usepackage{textcomp}
\usepackage{cite}
\usepackage{algorithm}
\usepackage{algpseudocode}
\usepackage{multicol}
\usepackage{multirow}

\title{Design of a Fully Actuated 4-DOF Robotic Finger With Joint-Specific Hybrid Remote Actuation}

\author{
    Hyojae Kang$^{1}$, Hyun-mok Jeong$^{1}$, Joonho Lee$^{1}$, Dongil Park$^{1,2}$, Hyunmin Do$^{1}$,\\ Jongwoo Park$^{1,*}$, and Jeongdo Ahn$^{1,*}$ \\
    $^{1}$Advanced Robotics Research Center, Korea Institute of Machinery \& Materials (KIMM) \\
    $^{2}$Mechanical Engineering, University of Science \& Technology (UST) \\
    $^{*}$Co-Corresponding author
}

\begin{document}
\date{}
\maketitle

\thispagestyle{plain}
\footnotetext{This manuscript has been submitted for possible publication.\\
This work was supported by the Industrial Strategic Technology Development Program (00443339) funded by the Ministry of Trade, Industry \& Energy(MOTIE, Korea).}

\begin{abstract}
This paper presents a fully actuated 4-DOF robotic finger using a joint-specific hybrid remote-actuation architecture. The metacarpophalangeal (MCP) joint is driven by two coordinated rigid-link transmission sets, whereas the proximal interphalangeal (PIP) and distal interphalangeal (DIP) joints are independently actuated by closed-loop wire transmissions incorporating circular rolling-contact joints (RCJs). A larger transmission radius is used at the PIP joint than at the DIP joint. The RCJ wire geometry maintains the total wire-loop length during joint rotation, and the DIP wire routing is designed so that PIP motion does not affect its differential actuation for a fixed MCP configuration. The distal wire transmission, MCP linkage, and fingertip kinematics are analytically modeled. In experiments, the transmission behavior was quantitatively evaluated from the ball-screw displacements measured using ArUco-marker tracking. MCP actuation produced measurable displacements of the PIP and DIP transmission units, whereas isolated PIP and DIP actuation with the MCP fixed supported the intended mechanical decoupling between the distal transmissions. The mean peak fingertip forces under isolated MCP, PIP, and DIP actuation were 21.28~N, 9.22~N, and 5.75~N, respectively. The resulting finger postures were also examined using objects of different geometries and sizes.
\end{abstract}

\section{Introduction}

Robotic hands require multiple controllable degrees of freedom (DoFs) to generate diverse finger postures for grasping and manipulation \cite{Bicchi2000}. Fully or highly actuated hands preserve joint-level controllability by assigning independent actuation to multiple finger DoFs \cite{Butterfass2001,Martin2014}. However, increasing the number of actuated joints also increases the number of actuators and transmission components. Packaging these components within the hand is especially restrictive in multi-DoF fingers because the available installation volume decreases toward the distal phalanges, while distal actuator placement increases the mass and structural volume carried by the finger.

Remote actuation separates actuator placement from joint location. Rigid-linkage transmission provides deterministic kinematics and direct force transmission, whereas tendon- or wire-driven transmission provides greater routing flexibility through confined structures \cite{Lovchik1999,Rothling2007Shadow,Kim2021ILDA,Gossen2025LoA}. Their complementary characteristics have motivated hybrid linkage--tendon architectures \cite{Yang2021MCR,Cerruti2016ALPHA,Zhang2018CATCH,Li2026REL}. However, the appropriate transmission type can differ according to joint location and mechanical function. For a proximal multi-DoF joint such as the metacarpophalangeal (MCP) joint, a rigid-link transmission can transmit the motion of multiple proximal actuators to a multi-DoF joint while avoiding additional tendon-routing elements around the finger base. In contrast, flexible transmission is advantageous for serial distal joints because it allows actuation to be remotely located without placing actuators within the phalanges.

Extending wire transmission to multiple serial joints introduces additional requirements. Bidirectional actuation requires either passive restoration or opposing agonist--antagonist transmission paths because a wire transmits tensile force only while tensioned \cite{Lee1994,Grebenstein2008,Grebenstein2010,Hu2022,Li2026SoftHandA,Byun2025LUTA}. In addition, a wire driving a distal joint must traverse moving proximal joints, and configuration-dependent changes in routing length can generate unintended interjoint coupling. Recent studies have addressed these issues using antagonistic remote transmission and joint-centered routing \cite{Choi2026Path,Min2025Bowden,Tajima2026}. A fully actuated multi-DoF finger therefore requires both independent actuation inputs and a transmission arrangement that accommodates the different mechanical requirements of the proximal and distal joints while avoiding unintended coupling between the distal joints.

This paper presents a fully actuated 4-DOF robotic finger based on a joint-specific hybrid remote-actuation architecture, combining rigid-link transmission for the 2-DOF MCP joint with independent closed-loop wire transmissions for the proximal interphalangeal (PIP) and distal interphalangeal (DIP) joints using circular rolling-contact joints (RCJs). This architecture retains independent actuation of all four DoFs without placing actuators in the distal finger links, while the distal wire routing prevents differential coupling between the PIP and DIP transmissions for a fixed MCP configuration. Experiments quantify the MCP-induced displacement of the distal transmission units and evaluate the mechanical decoupling between the PIP and DIP transmissions for a fixed MCP configuration, together with the fingertip force and object-interaction behavior of the fabricated finger.

The main contributions are:
\begin{itemize}
    \item A joint-specific hybrid remote-actuation architecture for a fully actuated 4-DOF finger, combining two coordinated rigid-link transmission sets at the 2-DOF MCP joint with remote wire transmissions at the serial PIP and DIP joints.
    \item A closed-loop RCJ wire transmission that maintains bidirectional actuation and provides mechanical decoupling between the PIP and DIP transmissions for a fixed MCP configuration.
\end{itemize}

The remainder of this paper is organized as follows. Section~II reviews robotic finger actuation architectures relevant to the proposed mechanism. Section~III presents the mechanical design, and Section~IV provides the kinematic and transmission analyses. Section~V experimentally evaluates joint motion, fingertip force, and representative object interaction. Section~VI concludes the paper.

\section{Related Works}

Robotic fingers and hands can be broadly categorized as underactuated or fully actuated according to the relationship between their DoFs and degrees of actuation (DoAs). Underactuated mechanisms use fewer actuators than controllable joint DoFs, typically relying on mechanical coupling, passive adaptation, or compliant elements to coordinate multiple joint motions \cite{Dollar2010SDM,Catalano2014SoftHand,Odhner2014iHY,Deimel2016RBO,Kontoudis2019,kang2026design}. In contrast, fully actuated systems including DLR-Hand II \cite{Butterfass2001}, KITECH-Hand \cite{Lee2017KITECH}, LEAP Hand \cite{Shaw2023LEAP}, RAPID Hand\cite{rapid}, and EyeSight Hand \cite{Romero2024EyeSight} assign independent actuation to each controlled DoF, enabling direct control of multiple finger coordinates. This increased controllability, however, requires a corresponding number of actuators and transmission components, making actuator packaging a major design consideration, particularly in multi-DoF fingers. Remote transmission addresses this issue by separating actuator placement from joint location. The ILDA Hand \cite{Kim2021ILDA} employs a rigid-linkage-based proximal transmission, while hybrid systems such as MCR-Hand II \cite{Yang2021MCR} and REL Hand \cite{Li2026REL} distribute linkage and tendon transmission among different finger functions.

Tendon- and wire-driven fingers enable actuators to be located away from distal joints, but their performance depends on tension management and routing geometry \cite{Lovchik1999,Grebenstein2010,Gossen2025LoA,Tajima2026}. Passive elastic restoration has been used in underactuated mechanisms such as the LUTA Hand \cite{Byun2025LUTA}, whereas antagonistic architectures actively drive both directions using opposing tensile paths \cite{Lee1994,Grebenstein2008,Hu2022,Li2026SoftHandA,Min2025Bowden}. For serially arranged joints, the wire path to a distal joint can also change as a proximal joint moves, producing unwanted coupling. Choi \textit{et al.} \cite{Choi2026Path} addressed this problem by routing wires through revolute-joint centers so that proximal rotation does not alter the effective routing length of distal transmissions. 

RCJs provide an alternative to conventional pin joints by constraining adjacent surfaces to roll relative to each other, allowing the joint trajectory and transmission geometry to be designed simultaneously \cite{Collins2003}. RCJs have been applied to underactuated prosthetic fingers \cite{Jobin2004}, tendon-driven compliant manipulators for cable-slack suppression \cite{Zhang2021CRCJ}, adaptive hands with passive elastic restoration \cite{Byun2025LUTA}, and mechanisms with optimized noncircular rolling surfaces and transmission geometries \cite{Decker2026NCRCJ}.

In the proposed finger, the proximal 2-DOF MCP joint is driven by two rigid-link transmission sets, whereas the serial PIP and DIP joints are driven by closed-loop wire transmissions. This joint-specific allocation avoids additional tendon-routing elements around the MCP while keeping the distal actuators outside the phalanges. At the PIP and DIP joints, circular RCJs are incorporated into closed agonist--antagonist wire loops for bidirectional actuation. The DIP wire is routed through the PIP structure so that PIP motion does not produce differential DIP actuation, providing mechanical decoupling between the two distal transmissions for a fixed MCP configuration. Thus, the architecture combines full joint-level actuation with remote distal actuation without requiring a single transmission type to satisfy the different mechanical requirements of the proximal and distal joints.

\section{Mechanical Design}

\begin{figure*}[!t]
    \centering
    \includegraphics[width=0.99\textwidth]{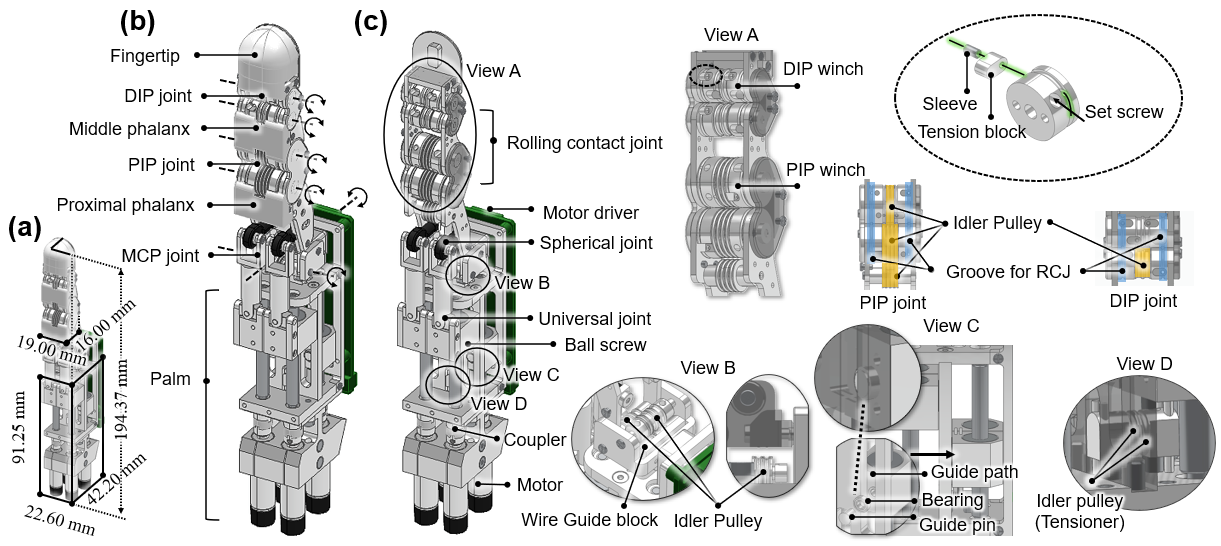}
    \caption{Design of the proposed finger: (a) Dimensions, (b) Overall configuration and joint arrangement, (c) Detailed transmission components.}
    \label{fig:mechanical_design}
\end{figure*}

The proposed robotic finger is driven by four motors and provides four independently actuated DoFs: abduction/adduction (A/A) and flexion/extension (F/E) at the MCP joint, and F/E at the PIP and DIP joints. Figure~\ref{fig:mechanical_design} shows the overall configuration, principal dimensions, and detailed transmission components of the fabricated finger.

As shown in Fig.~\ref{fig:mechanical_design}(a), the total module length is less than 200~mm, while the finger portion has a width and thickness of 19.00~mm and 16.09~mm, respectively. The major anatomical segments and joints are identified in Fig.~\ref{fig:mechanical_design}(b). From the distal end, the finger consists of the fingertip, DIP joint, middle phalanx, PIP joint, proximal phalanx, and MCP joint, followed by the palm-side actuation structure.

The MCP transmission adapts the proximal rigid-link actuation concept used in the ILDA Hand \cite{Kim2021ILDA} and incorporates two rigid-link transmission sets integrated with ball-screw actuation. In each set, a universal joint connected to the ball-screw transmission is mechanically linked to a spherical joint connected to the MCP structure. Coordinated motion of the two sets generates MCP A/A and F/E about the axes indicated in Fig.~\ref{fig:mechanical_design}(b). The two rigid-link transmission sets generate MCP A/A and F/E through coordinated ball-screw motion without adding antagonistic wire paths around the finger base, through which the PIP and DIP actuation wires also pass. 

The detailed distal transmission architecture is shown in Fig.~\ref{fig:mechanical_design}(c). View A illustrates the PIP and DIP RCJ assemblies and their corresponding wire winches. The PIP winch has a larger effective transmission radius than the DIP winch, providing a larger wire moment arm at the PIP joint. Each winch includes a circumferential groove for wire routing and a through-hole for wire termination. The wire is secured using a sleeve and a tension block, and its initial tension can be adjusted by driving the tension block with a set screw. The wire end is additionally fixed using stainless-steel solder and flux. The same termination structure is used for the wires that maintain the rolling-contact constraint of the RCJs. Idler pulleys positioned around the PIP and DIP joints guide the actuation wires toward the proximal transmission system.

The PIP and DIP RCJs are actuated through bidirectional closed-loop wire transmissions driven by ball screws. The wire fixed at each RCJ winch passes through the joint-level idler pulleys and then through the proximal routing components shown in View B, consisting of an idler pulley, a wire guide block, and an additional idler pulley, before reaching the block connected to the ball-screw nut. The wire ends are terminated by crimped sleeves at the nut block. One branch of the bidirectional wire is connected directly to the block, whereas the opposing branch passes through the adjustable idler pulley shown in View D. The position of this idler can be adjusted using a bolt connected to the lower frame, allowing the initial wire tension to be regulated. As shown in View C, a bearing mounted on a guide pin follows the guide path to constrain the nut block during ball-screw translation.

\begin{figure}[!t]
    \centering
    \includegraphics[width=0.99\columnwidth]{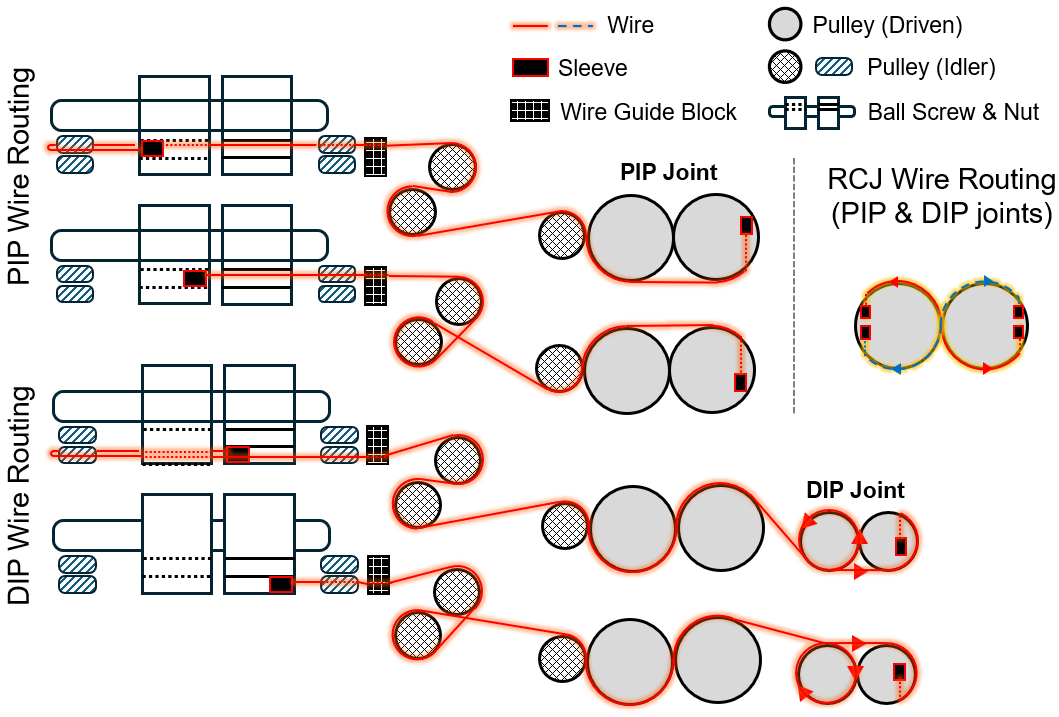}
    \caption{Wire-routing configuration of the finger.}
    \label{fig:wire_routing}
\end{figure}

Three wire-routing systems are required for the PIP and DIP mechanisms: the wires that maintain the rolling-contact constraint of the RCJs, the PIP actuation wire, and the DIP actuation wire. Their complete routing paths are schematically illustrated in Fig.~\ref{fig:wire_routing}. Only the components directly involved in wire transmission are retained in the schematic to clarify the routing topology. The actuation wires are 0.45 mm-diameter 7$\times$19 stranded cables made of SUS304 stainless steel. The RCJ constraint wires circulate around the paired rolling surfaces, whereas the PIP and DIP actuation wires are routed from the ball-screw-driven blocks through the corresponding idlers to their respective joint winches. With the MCP joint fixed, the PIP and DIP actuation wires are routed through separate differential paths. The DIP path passes through the PIP structure, while MCP motion changes the proximal routing lengths of both distal actuation wires because they pass through the moving MCP structure.

\section{Transmission and Kinematic Analysis}

\subsection{RCJ Wire Transmission}

\begin{figure}[!t]
    \centering
    \includegraphics[width=0.42\columnwidth]{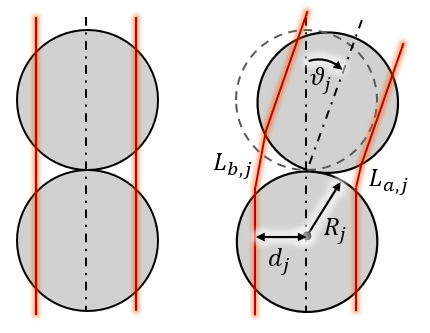}
    \caption{Equivalent RCJ actuation-wire geometry for closed-loop wire-length.}
    \label{fig:rcj_kinematics}
\end{figure}

For the adopted RCJ wire transmission, bidirectional actuation requires the total wire-loop length to remain unchanged during joint rotation. For the adopted RCJ geometry shown in Fig.~\ref{fig:rcj_kinematics}, the wire-hole height is equal to the rolling-surface radius. Accordingly, the agonist and antagonist wire lengths of the $j$th RCJ can be expressed as

\begin{equation}
L_{a,j}
=
2R_j+2d_j\sin\frac{\vartheta_j}{2},
\qquad
L_{b,j}
=
2R_j-2d_j\sin\frac{\vartheta_j}{2},
\label{eq:wire_length}
\end{equation}

\noindent where $j\in\{P,D\}$ denotes the PIP or DIP joint, $R_j$ is the rolling-surface radius, $d_j$ is the wire-hole offset, and $\vartheta_j$ denotes the relative rotation of the corresponding RCJ from its reference configuration. Therefore,

\begin{equation}
L_{a,j}+L_{b,j}=4R_j,
\label{eq:wire_invariance}
\end{equation}

\noindent which is independent of joint rotation. Thus, the increase in one wire branch is compensated by an equal decrease in the opposing branch, preserving the total wire-loop length during bidirectional motion.

The two contacting members of each RCJ rotate with a $1{:}1$ motion ratio. Accordingly, the relative rotations of the PIP and DIP joints are related to the corresponding generalized coordinates by

\begin{equation}
\vartheta_P=2q_3,
\qquad
\vartheta_D=2q_4.
\label{eq:rcj_ratio}
\end{equation}

The differential wire displacement associated with the $j$th RCJ is defined as

\begin{equation}
u_j
=
\frac{L_{a,j}-L_{b,j}}{2}
=
2d_j\sin\frac{\vartheta_j}{2}.
\label{eq:wire_displacement}
\end{equation}

Using Eq.~\eqref{eq:rcj_ratio}, the local actuation displacement generated at each RCJ becomes

\begin{equation}
u_P^{\mathrm{RCJ}}
=
2d_P\sin q_3,
\qquad
u_D^{\mathrm{RCJ}}
=
2d_D\sin q_4.
\label{eq:local_wire_displacement}
\end{equation}

The PIP actuation wire terminates at the PIP joint and does not traverse the DIP joint. Therefore, its differential actuation displacement is independent of the DIP configuration,

\begin{equation}
u_P(q_3,q_4)
=
u_P^{\mathrm{RCJ}}(q_3),
\qquad
\frac{\partial u_P}{\partial q_4}=0.
\label{eq:pip_decoupling}
\end{equation}

In contrast, the DIP actuation wire passes through the PIP structure before reaching the DIP joint. The total routed lengths of its agonist and antagonist branches can therefore be expressed as

\begin{align}
\Lambda_{a,D}
&=
C_{a,D}
+
L_{a,D}^{(P)}
+
L_{a,D}^{(D)}(q_4),
\\
\Lambda_{b,D}
&=
C_{b,D}
+
L_{b,D}^{(P)}
+
L_{b,D}^{(D)}(q_4),
\label{eq:dip_total_path}
\end{align}

\noindent where $C_{a,D}$ and $C_{b,D}$ contain the configuration-independent portions of the routing, $L_{a,D}^{(P)}$ and $L_{b,D}^{(P)}$ denote the wire lengths associated with the passage through the PIP structure, and $L_{a,D}^{(D)}$ and $L_{b,D}^{(D)}$ denote the local contributions at the DIP RCJ.

\begin{figure}
    \centering
    \includegraphics[width=0.99\columnwidth]{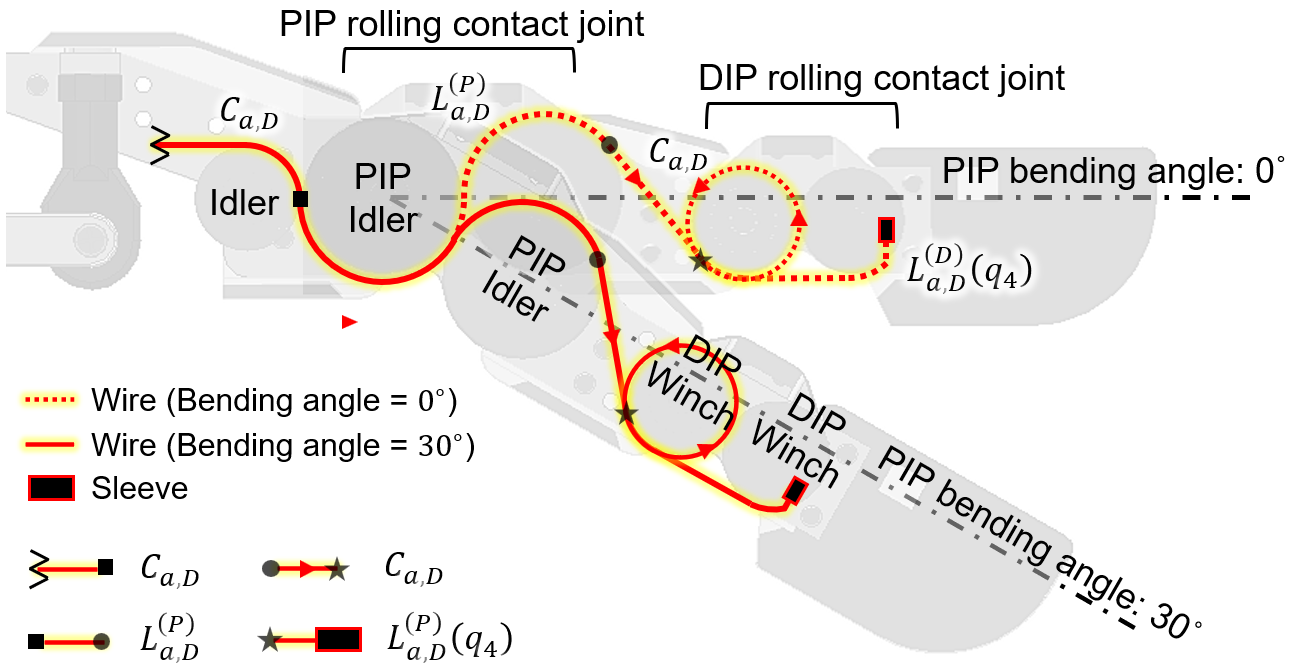}
    \caption{DIP wire routing through the PIP RCJ during PIP rotation.}
    \label{fig:driving_wire}
\end{figure}

The DIP wire is routed through the PIP structure such that the length of the PIP-passage segment remains unchanged during PIP rotation as shown in the Fig.~\ref{fig:driving_wire}. Furthermore, the corresponding passage lengths of the two opposing DIP actuation branches are identical. Therefore,

\begin{equation}
L_{a,D}^{(P)}
=
L_{b,D}^{(P)}
=
L_D^{(P)}
=
\mathrm{const.},
\label{eq:dip_common_mode}
\end{equation}

\noindent where $L_D^{(P)}$ denotes the constant wire length associated with the passage through the PIP structure. Thus, PIP rotation does not introduce any configuration-dependent variation into the DIP actuation-wire length through this passage. When the differential DIP actuation displacement is formed, the identical constant PIP-passage terms cancel between the agonist and antagonist branches. After absorbing the remaining constant branch-length difference into the reference configuration and initial pretension, the effective DIP displacement for a fixed MCP configuration becomes

\begin{equation}
u_D(q_3,q_4)
=
u_D^{\mathrm{RCJ}}(q_4)
=
2d_D\sin q_4,
\qquad
\frac{\partial u_D}{\partial q_3}=0.
\label{eq:dip_decoupling}
\end{equation}

Thus, this geometric property provides intrinsic mechanical decoupling of the DIP transmission from PIP motion. Collecting the PIP and DIP actuation displacements gives

\begin{equation}
\mathbf{u}
=
\begin{bmatrix}
u_P\\
u_D
\end{bmatrix}
=
\begin{bmatrix}
2d_P\sin q_3\\
2d_D\sin q_4
\end{bmatrix}.
\label{eq:wire_displacement_vector}
\end{equation}

The corresponding differential transmission relation is

\begin{equation}
\dot{\mathbf{u}}
=
\mathbf{J}_{w}
\dot{\mathbf{q}}_{d},
\qquad
\mathbf{q}_{d}
=
\begin{bmatrix}
q_3\\
q_4
\end{bmatrix},
\label{eq:wire_velocity}
\end{equation}

\noindent where

\begin{equation}
\mathbf{J}_{w}
=
\frac{\partial\mathbf{u}}
{\partial\mathbf{q}_{d}}
=
\begin{bmatrix}
2d_P\cos q_3 & 0\\
0 & 2d_D\cos q_4
\end{bmatrix}.
\label{eq:wire_jacobian}
\end{equation}

The zero off-diagonal terms in Eq.~\eqref{eq:wire_jacobian} indicate that, for the ideal routing geometry and a fixed MCP configuration, PIP motion does not generate a differential DIP actuation displacement and DIP motion does not generate a PIP actuation displacement. This diagonal transmission relation represents the intended mechanical decoupling between the PIP and DIP wire transmissions.

\subsection{MCP Rigid-Link Transmission}

The geometric relationship between the two ball-screw inputs and the MCP joint configuration is described using the rigid-link transmission model shown in Fig.~\ref{fig:kinematic_model}. Let $O$ denote the finger-base origin, and let $P_1$ and $P_2$ denote the ball-screw-side universal-joint centers of the two MCP transmission sets. The corresponding ball-screw displacements are denoted by $d_1$ and $d_2$, respectively. The MCP joint center is denoted by $P_5$, while $P_3$ and $P_4$ denote the centers of the spherical joints connected to the MCP structure. The coordinate vector of point $P_i$ with respect to the origin $O$ is denoted by $\mathbf{p}_i$, whereas $\mathbf{r}_{ij}$ denotes the vector from $P_i$ to $P_j$. The coordinate of $P_3$ is obtained from the MCP-joint position and the rotated offset vector as

\begin{equation}
\mathbf{p}_3
=
\mathbf{p}_5
+
\mathbf{Rot}(q_1,q_2)\mathbf{r}_{53},
\label{eq:mcp_p3}
\end{equation}

\noindent where $q_1$ and $q_2$ denote MCP A/A and F/E, respectively. From the geometry shown in Fig.~\ref{fig:kinematic_model}, the MCP-joint coordinate is

\begin{equation}
\mathbf{p}_5
=
\begin{bmatrix}
0 & 0 & d_a
\end{bmatrix}^{T},
\label{eq:mcp_p5}
\end{equation}

\noindent and the fixed offset vector from $P_5$ to $P_3$ is defined as

\begin{equation}
\mathbf{r}_{53}
=
\begin{bmatrix}
l_{u2}\cos\beta \\
l_{u2}\sin\beta\cos\alpha \\
l_{u2}\sin\beta\sin\alpha
\end{bmatrix}.
\label{eq:mcp_r53}
\end{equation}

The corresponding coordinate $\mathbf{p}_4$ can be obtained in the same manner using the fixed offset vector $\mathbf{r}_{54}$. The positions of the ball-screw-side universal joints vary along their respective actuation axes. Accordingly, their coordinate vectors can be written as

\begin{equation}
\mathbf{p}_1
=
\mathbf{p}_{1,0}
+
d_1\mathbf{e}_1,
\qquad
\mathbf{p}_2
=
\mathbf{p}_{2,0}
+
d_2\mathbf{e}_2,
\label{eq:mcp_ball_screw}
\end{equation}

\noindent where $\mathbf{p}_{1,0}$ and $\mathbf{p}_{2,0}$ are the coordinates at the reference configuration, and $\mathbf{e}_1$ and $\mathbf{e}_2$ are unit vectors along the corresponding ball-screw axes. Because the rigid links connecting $P_1$ to $P_3$ and $P_2$ to $P_4$ maintain the constant length $l_{u1}$, the transmission constraints are expressed as

\begin{equation}
\left\|
\mathbf{p}_3-\mathbf{p}_1
\right\|
=
l_{u1},
\qquad
\left\|
\mathbf{p}_4-\mathbf{p}_2
\right\|
=
l_{u1}.
\label{eq:mcp_link_constraint}
\end{equation}

Therefore, for given MCP joint angles $q_1$ and $q_2$, the positions of $P_3$ and $P_4$ are first determined from the MCP geometry, and the corresponding ball-screw displacements $d_1$ and $d_2$ can then be obtained from Eq.~\eqref{eq:mcp_link_constraint}. The geometric parameters required for the MCP transmission model are summarized in Table~\ref{tab:kinematic_parameters}.

\subsection{Finger Kinematics}

The finger configuration is represented by

\begin{equation}
\mathbf{q}
=
\begin{bmatrix}
q_1 & q_2 & q_3 & q_4
\end{bmatrix}^{T},
\label{eq:joint_vector}
\end{equation}

\noindent where $q_1$ and $q_2$ represent MCP A/A and F/E, respectively, while $q_3$ and $q_4$ represent the rolling coordinates associated with the PIP and DIP RCJs.

\begin{figure}[!t]
    \centering
    \includegraphics[width=0.99\columnwidth]{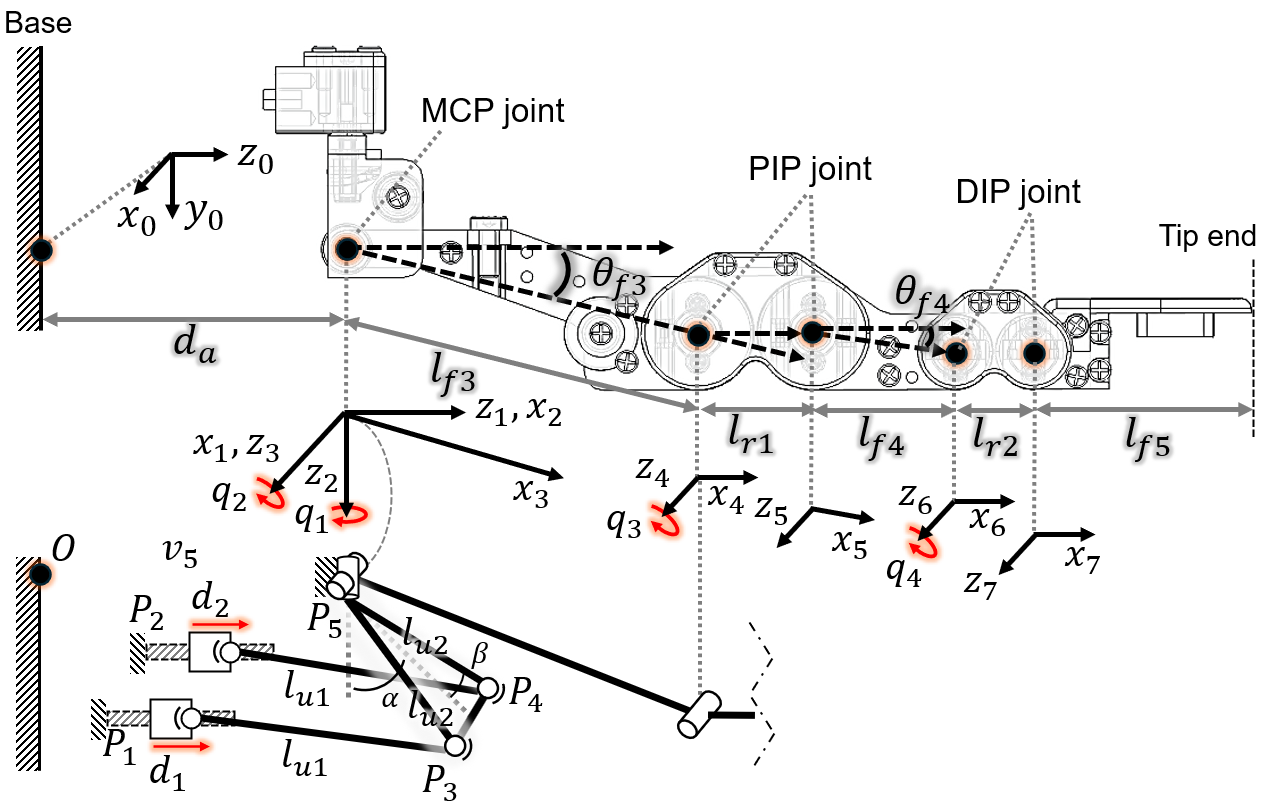}
    \caption{Coordinate frames of the finger and kinematic diagram of the MCP joint actuation.}
    \label{fig:kinematic_model}
\end{figure}

The coordinate frames used for the kinematic formulation are illustrated in Fig.~\ref{fig:kinematic_model}. The forward kinematics are formulated using Craig's modified Denavit--Hartenberg (MDH) parameters summarized in Table~\ref{tab:mdh_parameter}.

\begin{table}[h]
\centering
\caption{MDH parameters of the proposed finger}
\label{tab:mdh_parameter}
\renewcommand{\arraystretch}{1.15}
\begin{tabular}{c c c c c}
\hline
$i$ & $a_{i-1}$ [mm] & $\alpha_{i-1}$ [deg] & $d_i$ [mm] & $\theta_i$ \\
\hline
1 & 0        & 0     & $d_a$ & 0 \\
2 & 0        & $-90$ & 0 & $q_1 - 90$ \\
3 & 0        & $-90$ & 0 & $q_2 - \theta_{f3}$ \\
4 & $l_{f3}$ & 0     & 0 & $q_3 + \theta_{f3}$ \\
5 & $l_{r1}$ & 0     & 0 & $q_3 - \theta_{f4}$ \\
6 & $l_{f4}$ & 0     & 0 & $q_4 + \theta_{f4}$ \\
7 & $l_{r2}$ & 0     & 0 & $q_4$ \\
\hline
\end{tabular}
\end{table}

The homogeneous transformation associated with the $i$th row of Table~\ref{tab:mdh_parameter} is defined as

\begin{equation}
{}^{i-1}\mathbf{A}_{i}
=
\mathbf{R}_{x}(\alpha_{i-1})
\mathbf{P}_{x}(a_{i-1})
\mathbf{R}_{z}(\theta_i)
\mathbf{P}_{z}(d_i),
\label{eq:mdh_transform}
\end{equation}

\noindent where $\mathbf{R}_{x}$ and $\mathbf{R}_{z}$ denote rotations about the corresponding axes, and $\mathbf{P}_{x}$ and $\mathbf{P}_{z}$ denote translations.

Using the MDH parameters, the transformation from the finger base frame to the fingertip is given by

\begin{equation}
{}^{0}\mathbf{T}_{\mathrm{tip}}
=
\left(
\prod_{i=1}^{7}
{}^{i-1}\mathbf{A}_{i}
\right)
\mathbf{P}_{x}(l_{f5}).
\label{eq:fingertip_fk}
\end{equation}

The repeated appearances of $q_3$ and $q_4$ in Table~\ref{tab:mdh_parameter} represent the rotations of the two contacting members of each RCJ. Consequently, the relative PIP and DIP joint rotations are $2q_3$ and $2q_4$, respectively, consistent with Eq.~\ref{eq:rcj_ratio}.

The fingertip position is obtained from the translational component of Eq.~\eqref{eq:fingertip_fk} as

\begin{equation}
\mathbf{p}_{\mathrm{tip}}
=
\begin{bmatrix}
x & y & z
\end{bmatrix}^{T}
=
\mathrm{trans}
\left(
{}^{0}\mathbf{T}_{\mathrm{tip}}
\right).
\label{eq:fingertip_position}
\end{equation}

The ranges of the generalized coordinates used in the kinematic analysis are summarized in Table~\ref{tab:joint_rom}. Because of the $1{:}1$ rolling relation in Eq.~\eqref{eq:rcj_ratio}, the $45^\circ$ ranges of $q_3$ and $q_4$ correspond to $90^\circ$ relative bending ranges of the PIP and DIP RCJs.

\begin{table}[h]
\centering
\caption{Operating ranges of motion of the finger}
\label{tab:joint_rom}
\renewcommand{\arraystretch}{1.15}
\begin{tabular}{c c c}
\hline
Variable & Associated motion & Range [deg] \\
\hline
$q_1$ & MCP A/A & $-30$ to $30$ \\
$q_2$ & MCP F/E & $-78.40$ to $0$ \\
$q_3$ & PIP RCJ rolling motion & $-45$ to $0$ \\
$q_4$ & DIP RCJ rolling motion & $-45$ to $0$ \\
\hline
\end{tabular}
\end{table}

\begin{figure}[!t]
    \centering
    \includegraphics[width=0.99\columnwidth]{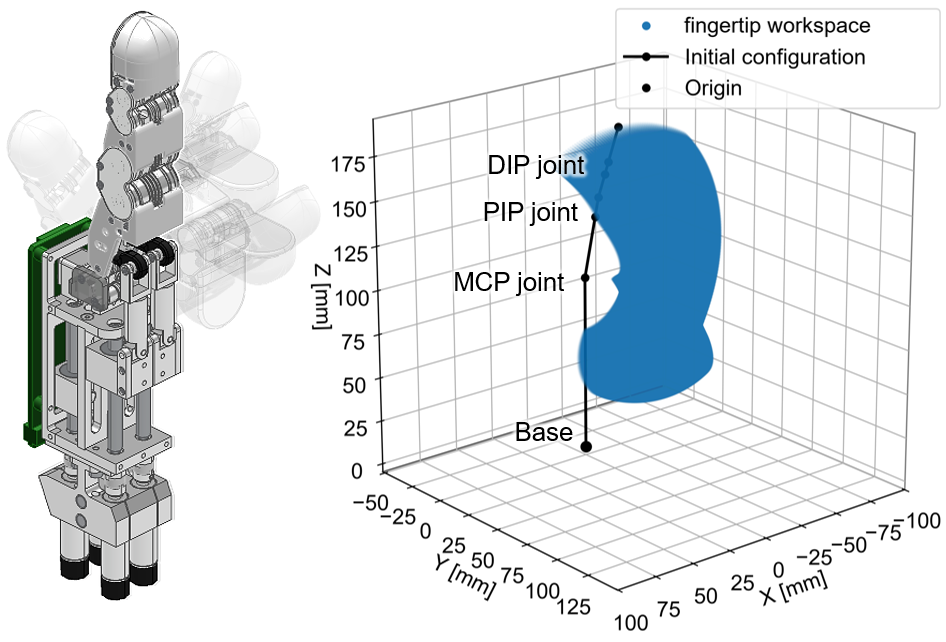}
    \caption{Reachable fingertip workspace of the proposed finger.}
    \label{fig:workspace_finger}
\end{figure}

Based on these coordinate ranges, the reachable workspace was calculated by evaluating Eq.~\eqref{eq:fingertip_position} over the joint space. Figure~\ref{fig:workspace_finger} shows the workspace of the proposed finger. The kinematic parameters used in the calculation are summarized in Table~\ref{tab:kinematic_parameters}.

\begin{table*}[h]
\centering
\caption{Kinematic Parameters of the Finger}
\label{tab:kinematic_parameters}
\begin{tabular}{c l c}
\hline
\textbf{Symbol} & \textbf{Description} & \textbf{Value} \\
\hline
$d_a$
& D$^1$ from RC$^2$ to the MCP joint
& 98.62 mm \\
$l_{u1}$
& L$^3$ of the MCP transmission link
& 28.01 mm \\
$l_{u2}$
& D$^1$ from the MCP joint to S$^4$
& 19.27 mm \\
$\mathbf{p}_{1,0}$
& RC$^2$ of the ball-screw-side U$^5$
& 6.39, 12.10, 85.98 mm \\
$\mathbf{p}_{2,0}$
& RC$^2$ of the ball-screw-side U$^5$
& -6.39, 12.10, 85.98 mm \\
$\alpha$
& IA$^6$ of the MCP-side S$^4$ offset
& 51.25$^\circ$ \\
$\beta$
& IA$^6$ of the MCP-side S$^4$ offset
& 19.67$^\circ$ \\
$l_{f3}$
& L$^3$ of the proximal finger link
& 38.07 mm \\
$l_{r1}$
& Link L$^3$ of the PIP RCJ
& 12.20 mm \\
$l_{f4}$
& L$^3$ of the middle finger link
& 15.13 mm \\
$l_{r2}$
& Link L$^3$ of the DIP RCJ
& 8.20 mm \\
$l_{f5}$
& L$^3$ of the distal finger link
& 23.00 mm \\
$\theta_{f3}$
& FRO$^7$ preceding the PIP joint
& $-13.67^\circ$ \\
$\theta_{f4}$
& FRO$^7$ in the PIP RCJ
& $-7.59^\circ$ \\
$R_P$
& R$^8$ of the PIP RCJ
& 6.00 mm \\
$R_D$
& R$^8$ of the DIP RCJ
& 4.00 mm \\
\hline
\multicolumn{3}{l}{D$^1$: Distance, RC$^2$: Reference Coordinate, L$^3$: Length,}\\
\multicolumn{3}{l}{S$^4$: Spherical Joint,
U$^5$: Universal Joint, IA$^6$: Installation Angle,}\\
\multicolumn{3}{l}{FRO$^7$: Fixed Rotational Offset, and R$^8$: Radius}
\end{tabular}
\end{table*}

\section{Experiments}

Experiments were conducted to evaluate the transmission behavior predicted by the proposed architecture and to characterize the fingertip force and object-interaction motion of the fabricated finger. The transmission tests focused on two behaviors: MCP-induced changes in the distal wire paths and mechanical decoupling between the PIP and DIP transmissions for a fixed MCP configuration. The main specifications of the finger are summarized in Table~\ref{tab:specification}. Videos of the tests described in Section~V-A, the backdrivability test, and the motion demonstration are provided in the supplemental material.

\begin{table*}[h]
\centering
\caption{Technical specifications of the finger.}
\label{tab:specification}
\begin{tabular}{l l}
\hline
\textbf{Specification} & \textbf{Proposed finger}\\
\hline
Degrees of freedom & 4 \\
Degrees of actuation & 4 \\
Motion & MCP A/A \& F/E, PIP F/E, DIP F/E \\
Fingertip force & 21.28 N (MCP Only) \\
                                      & 9.22 N (PIP Only) \\
                                      & 5.75 N (DIP Only) \\
Backdrivability & Demonstrated (Supplemental video) \\
Weight &  143.6 g\\
Power supply & 12V DC\\
Communication & CAN \\
Motor part number & DCX08M EB KL 12V (Maxon) \\
Gearhead & GPX08 A 16 to 1 (Maxon) \\
\hline
\end{tabular}
\end{table*}

\subsection{Motion and Transmission Evaluation}

The transmission behavior was evaluated from the displacements of the ball-screw units associated with each joint. During each transmission test, only the motor or motors required to generate the target joint motion were powered, while the remaining motors were unpowered so that their transmission units remained backdrivable. Two tests were conducted. In the first test, the distal wire displacement induced by MCP motion was evaluated from the displacements of the PIP and DIP ball-screw units. In the second test, the displacement responses of the MCP and DIP ball-screw units during PIP motion, and those of the MCP and PIP ball-screw units during DIP motion, were examined to verify the transmission decoupling. The displacements were measured by attaching ArUco markers to each ball-screw unit and tracking them from recorded video. The camera was calibrated using a ChArUco board (CC200-15-11.25) prior to data analysis.

\begin{figure}[!t]
 \centering
 \includegraphics[width=0.99\columnwidth]{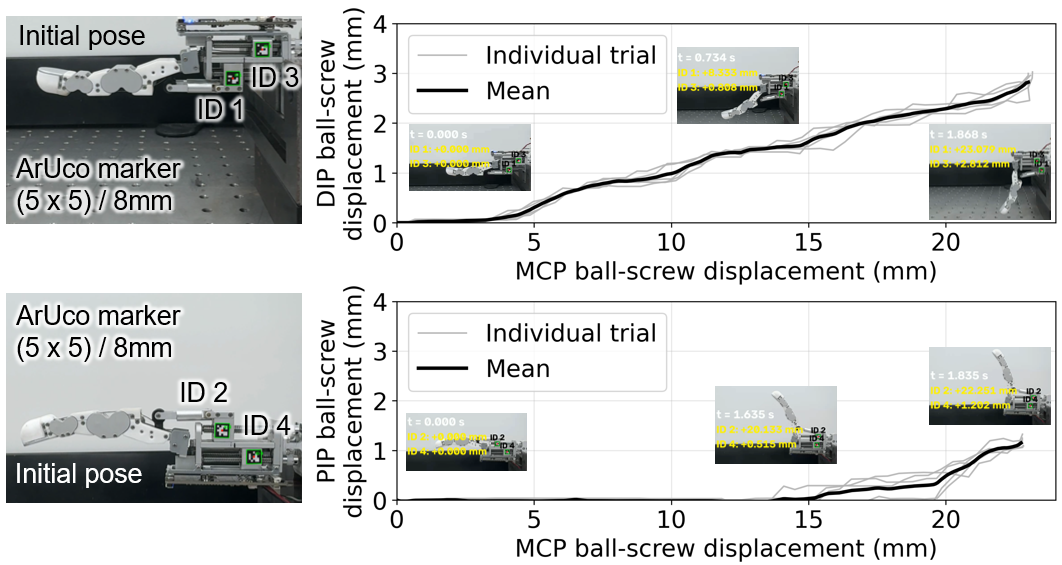}
 \caption{ArUco marker setup and measured displacement changes of the PIP and DIP ball-screw units induced by MCP motion.}
 \label{fig:aruco}
\end{figure}

Fig.~\ref{fig:aruco} shows the ArUco marker setup, the marker IDs, and the experimental results. The marker IDs were assigned from 1 to 4: IDs 1 and 2 correspond to the MCP ball-screw units, ID 3 to the DIP ball-screw unit, and ID 4 to the PIP ball-screw unit. The two plots show the displacement variations of the DIP and PIP ball-screw units, respectively, with respect to the driving displacement of the MCP ball-screw unit. The experiment was repeated five times, and the videos were recorded at 30 fps and analyzed from the onset of actual finger motion to its complete stop. At the maximum displacement of the MCP ball-screw unit, the average displacements of the DIP and PIP ball-screw units were 2.931 mm and 1.197 mm, respectively. In the CAD model, the corresponding MCP motion produces an ideal geometric routing-length change of 3.84~mm. However, in the fabricated system, nonideal factors such as wire tension and elastic deformation cause differences in the onset of actual transmission motion, even when MCP motion pulls the transmission paths of the other joints.

In the second test, no visible ball-screw motion was observed, consistent with the decoupled transmission relation derived in Section~IV. Although the ArUco-based measurements exhibited apparent displacement fluctuations of up to approximately 0.2 mm, the results support the intended mechanical decoupling between the PIP and DIP transmissions.

\begin{figure}
    \centering
    \includegraphics[width=0.99\columnwidth]{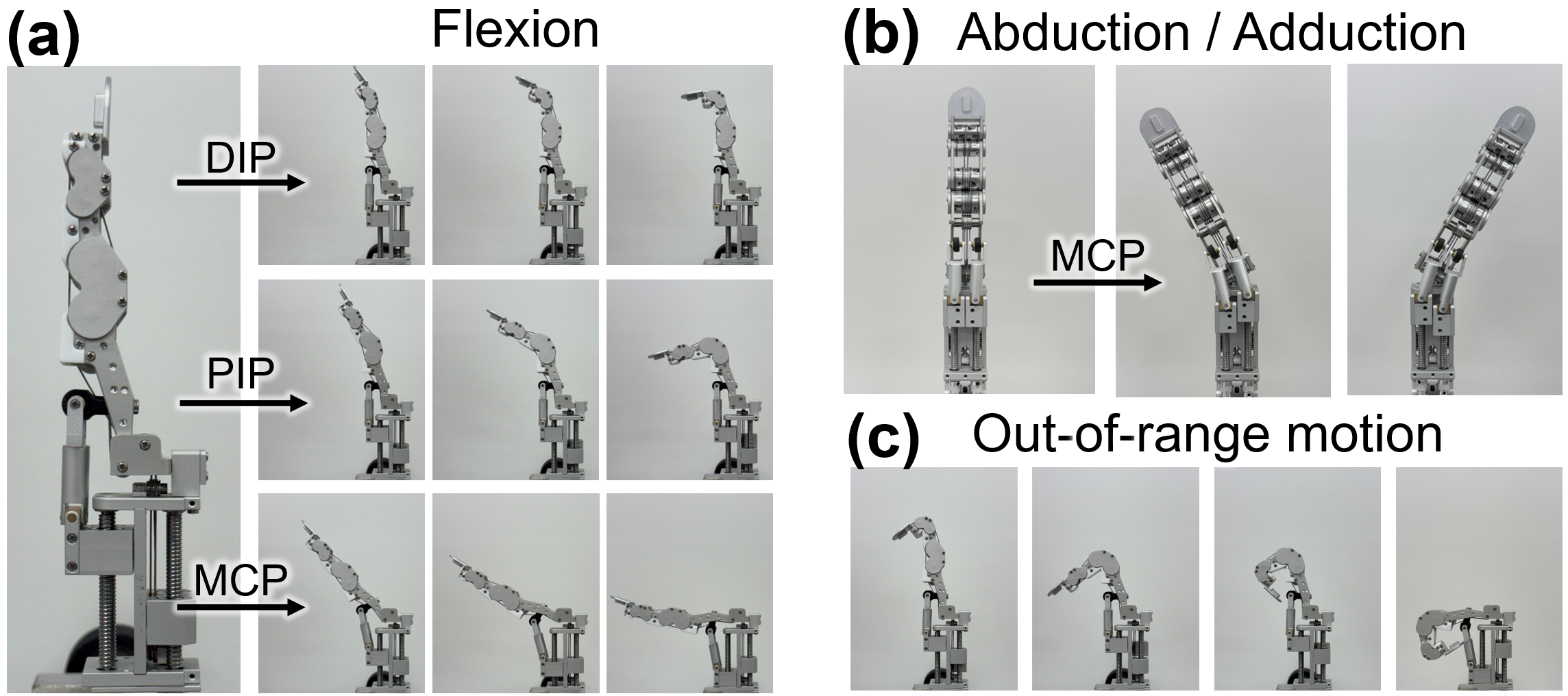}
    \caption{Representative motions of the fabricated finger: (a) F/E motion, (b) MCP A/A motion, and (c) motion beyond the nominal operating range.}
    \label{fig:motion_test}
\end{figure}

Figure~\ref{fig:motion_test} shows the representative motions generated by the fabricated finger. The F/E motions of the MCP, PIP, and DIP joints are shown in Fig.~\ref{fig:motion_test}(a), while the MCP A/A motion is shown in Fig.~\ref{fig:motion_test}(b). The joint ranges listed in Table~\ref{tab:joint_rom} represent the operating ranges adopted for the finger rather than the mechanical limits of the individual joints. The RCJs can rotate beyond these operating ranges where permitted by the surrounding finger geometry. Figure~\ref{fig:motion_test}(c) shows an example beyond the nominal operating range.

\subsection{Fingertip Force Measurement}

The fingertip force produced by isolated MCP, PIP, and DIP actuation was measured separately. A compression/tension load cell (QLMH-25, BIRAN) equipped with a hemispherical cap was used for the measurement. The fingertip was positioned above the hemispherical cap, and each joint was actuated independently to press the fingertip against the cap. The load-cell readings were converted to force using $g=9.812~\mathrm{m/s^2}$. The data were sampled at approximately 80 Hz, and the experiment was repeated three times for each joint.

\begin{figure}[!t]
    \centering
    \includegraphics[width=0.99\columnwidth]{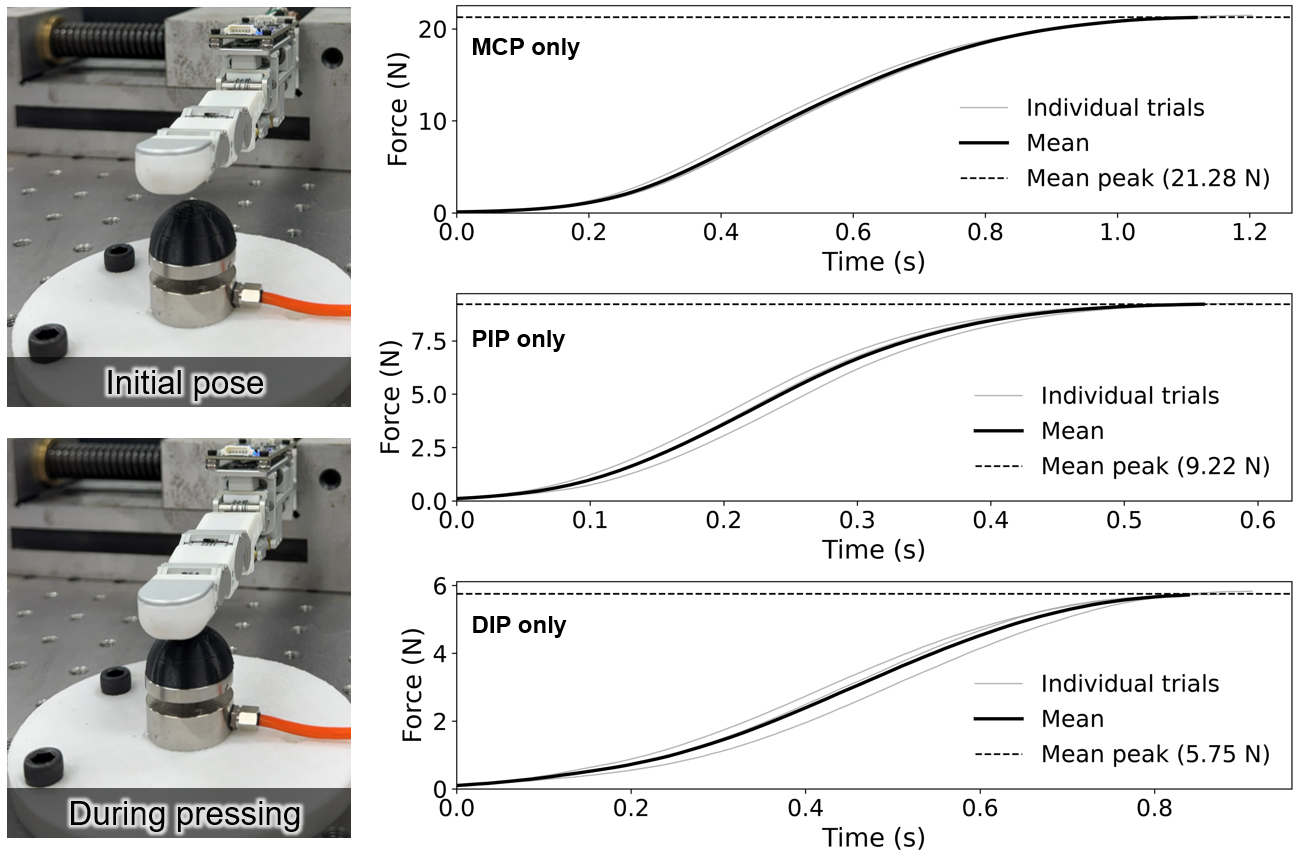}
    \caption{Experimental setup and fingertip force under independent actuation of each joint.}
    \label{fig:tipforce}
\end{figure}

Figure~\ref{fig:tipforce} shows the measured force profiles from the three trials and their mean. The dashed line denotes the mean peak force. The mean peak forces were 21.28 N, 9.22 N, and 5.75 N for isolated MCP, PIP, and DIP actuation, respectively. The fingertip force generated by MCP actuation depends on the finger configuration because its rigid-link transmission has a configuration-dependent mechanical advantage.

\subsection{Object Interaction}

\begin{figure}[!t]
    \centering
    \includegraphics[width=0.99\columnwidth]{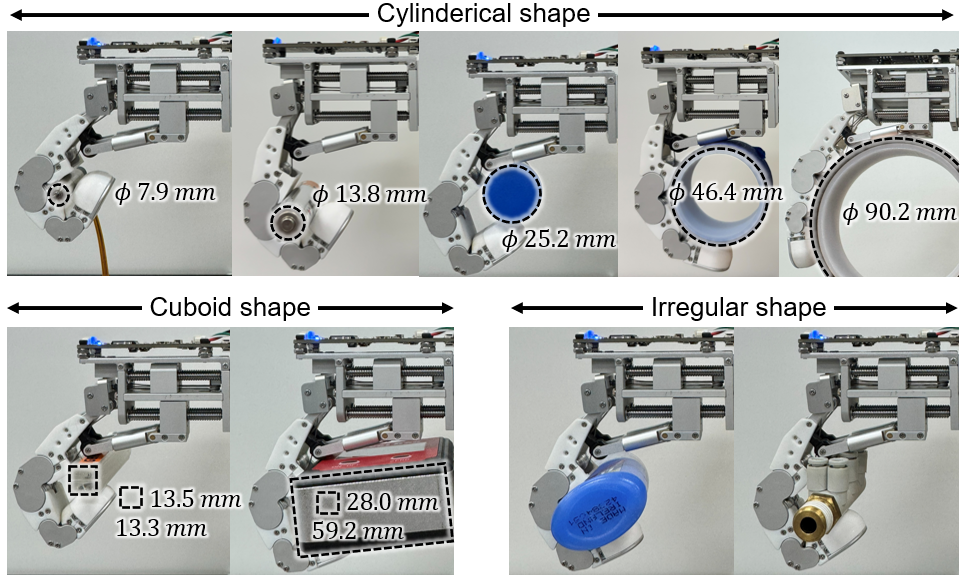}
    \caption{Finger postures during interaction with various objects.}
    \label{fig:gripping}
\end{figure}

Object-interaction tests were performed using five cylindrical objects of different diameters, two cuboid objects, and two irregularly shaped objects. For each object, the MCP, PIP, and DIP joints were adjusted to establish contact with the object. Fig.~\ref{fig:gripping} shows the resulting object-interaction postures. For the cylindrical objects, the joint configurations were adjusted according to the object diameter, while different configurations were used for the cuboid and irregular objects according to their external geometries.

\section{Conclusion}

This work presented and experimentally validated a joint-specific hybrid remote-actuation architecture for a fully actuated 4-DOF robotic finger. Rather than relying on a single transmission principle for all joints, the proposed design employs rigid-link transmission for the proximal 2-DOF MCP joint and closed-loop wire transmission with rolling-contact joints for the serial PIP and DIP joints. The analytical results showed that the RCJ-based wire transmission preserves the total agonist--antagonist wire-loop length during joint rotation and provides intrinsic mechanical decoupling between the PIP and DIP transmissions for a fixed MCP configuration.

Experiments with the fabricated finger confirmed the intended transmission behavior. ArUco-based displacement measurements showed that MCP actuation produced measurable displacement of the PIP and DIP transmission units, whereas isolated PIP and DIP actuation with the MCP configuration fixed produced no appreciable cross-displacement between the distal transmissions. In addition, the finger generated mean peak fingertip forces of 21.28~N, 9.22~N, and 5.75~N under isolated MCP, PIP, and DIP actuation, respectively, and achieved different contact postures for objects with various geometries. These results indicate that the proposed architecture can preserve independent joint-level actuation while relocating the distal actuators away from the finger links, providing a practical basis for constructing fully actuated dexterous robotic hands.

Future work will focus on reducing friction and other transmission losses in the wire-driven system and quantitatively characterizing transmission efficiency, hysteresis, and backdrivability. The proposed finger module will also be extended to a complete five-finger robotic hand, including a thumb. The resulting hand will be evaluated through systematic grasping experiments with objects of various geometries, followed by experimental assessment of its object-grasping and tool-manipulation capabilities.

\bibliographystyle{IEEEtran}
\bibliography{bibtex}

\end{document}